\documentclass[10pt,twocolumn,letterpaper]{article}

\pdfoutput=1

\usepackage{iccv}
\usepackage{times}
\usepackage{epsfig}
\usepackage{graphicx}
\usepackage{graphics}
\usepackage{amsmath}
\usepackage{amssymb}

\usepackage{multirow}
\usepackage{xcolor}
\usepackage{lipsum}
\usepackage{pdfpages}
\usepackage{float}
\usepackage{enumitem}

\newcommand{\prpl}{\textcolor[rgb]{0.59,0,0.67}}
\newcommand{\orng}{\textcolor[rgb]{1,0.63,0}}
\newcommand{\grn}{\textcolor[rgb]{0,1,0}}

\usepackage[pagebackref=true,breaklinks=true,letterpaper=true,colorlinks,bookmarks=false]{hyperref}

\iccvfinalcopy 

\def\iccvPaperID{0011} 

\ificcvfinal\fi

\begin{document}

\title{Weakly-Supervised \textit{Completion Moment} Detection using Temporal Attention}

\author{Farnoosh Heidarivincheh, Majid Mirmehdi and Dima Damen\\
Department of Computer Science\\
University of Bristol, Bristol, UK\\
{\tt\small \{farnoosh.heidarivincheh, M.Mirmehdi, Dima.Damen\}@bristol.ac.uk}
}

\maketitle
\ificcvfinal\thispagestyle{empty}\fi

\begin{abstract}
Monitoring the progression of an action towards completion offers fine grained insight into the actor's behaviour.
In this work, we target detecting the \textit{completion moment} of actions, that is the moment when the action's goal has been successfully accomplished. This has potential applications from surveillance to assistive living and human-robot interactions. Previous effort~\cite{Heidari2018} required human annotations of the \textit{completion moment} for training (i.e. full supervision). In this work, we present an approach for \textit{moment} detection from weak video-level labels. Given both \textit{complete} and \textit{incomplete} sequences, of the same action, we learn temporal attention, along with accumulated completion prediction from all frames in the sequence. 
We also demonstrate how the approach can be used when \textit{completion moment} supervision is available.
We evaluate and compare our approach on actions from three datasets, namely HMDB, UCF101 and RGBD-AC, and show that temporal attention improves detection in both weakly-supervised and fully-supervised settings.
\end{abstract}

\section{Introduction}
\label{sec:intro}
In the past few decades, with an increasing ubiquity and accessibility of video records, a significant part of research has been devoted to analysing human behaviour in video, including analysing human actions. This active area of research has vast applications, such as health-care, surveillance, video retrieval, entertainment, robotics and human-computer interaction. 
At its heart, action recognition has evolved from traditional hand-crafted features~\cite{laptev2005space,wang2013action,wang2013motionlets} to deep learning based approaches \cite{karpathy2014large,simonyan2014two,carreira2017quo,girdhar2017actionvlad} and achieved remarkable results. However, recent works have focused on proposing network architectures to deal with the spatio-temporal input, and neglected to explore fake or incomplete action instances. 

In this paper, we focus on incomplete actions, whether intentional or accidental, which could be crucial in contexts such as surveillance and health-care applications. These are actions which are attempted but their goals remain incomplete. 
Such incomplete sequences could be incorrectly recognised, or localised, by current state-of-the-art methods.
As an example, consider an \textit{incomplete pick}, where the subject only pretends to pick an object up. Standard action recognition classifiers would identify this as a \textit{pick} action, because of its similar motion to successfully completed picks. Likewise, a patient picking up a medicine tablet, but not ingesting it, would also be incorrectly recognised as ``take medicine'' action, posing risks to automatic monitoring of their health.

Action completion was first introduced in~\cite{Heidari2016} to assess whether the action's goal is achieved. The approach outputs sequence-level predictions of completion to distinguish \textit{complete} sequences from \textit{incomplete} ones. Subsequent works~\cite{Heidari2018,becattini2017am} proposed finer-grained analysis of an action's progression towards completion. For example, ~\cite{Heidari2018} looks for visual clues which confirm the goal's completion and detects the \textit{completion moment} from frame-level \textit{pre-} and \textit{post-completion} labels.

\begin{figure*}[h]
\begin{center}
\begin{tabular}{cc}
\includegraphics[width=.48\textwidth]{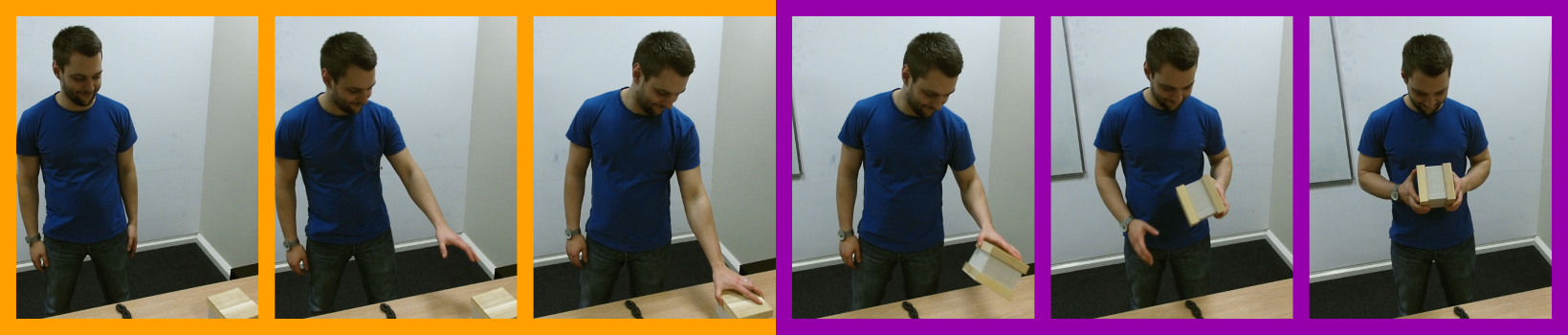}
&\includegraphics[width=.48\textwidth]{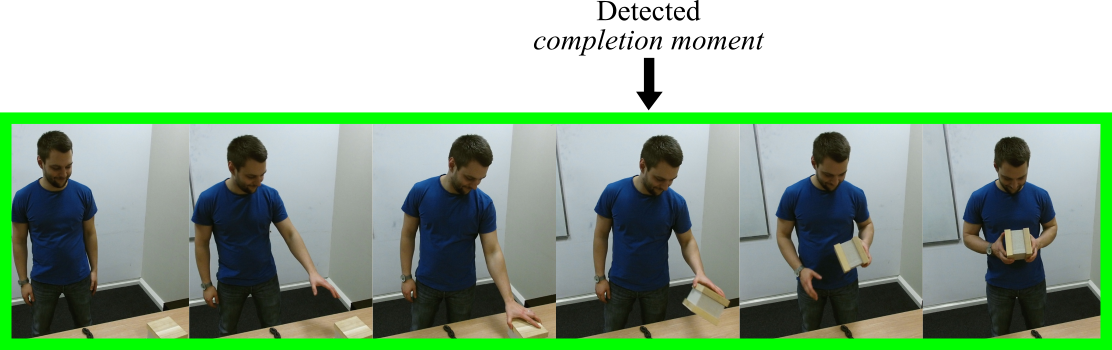}\\
Frame-level \orng{\textit{pre-}} and \prpl{\textit{post-completion}} labels & Sequence-level \grn{\textit{complete}} label
\end{tabular}
\caption{While frame-level labels (left) are used in a fully-supervised approach like~\cite{Heidari2016}, we detect the \textit{completion moment} only using weak labels, i.e. sequence level \textit{complete} and \textit{incomplete} (right).}
\label{fig:intro_labels}
\end{center}
\end{figure*}

In this work, we also investigate \textit{completion moment} detection. However, we differ from~\cite{Heidari2018} in the supervision by which our method learns to detect completion.
Frame-level annotations are not only expensive to collect, but importantly, highly subjective and often noisy~\cite{moltisanti2017trespassing,sigurdsson2016much,moltisanti19action}. We offer the first attempt to \textit{completion moment} detection with weak supervision, i.e. using only sequence-level \textit{complete} and \textit{incomplete} labels. Fig~\ref{fig:intro_labels} illustrates frame-level and sequence-level labels for a \textit{complete pick} action.
Given weak labels, we show that \textit{completion moment} detection could be achieved, by learning temporal attention.

We propose to use convolutional and recurrent cells with learnt temporal attention, and accumulate evidence for completion from all frames along the sequence, where evidence is weighted according to the frame's importance to the \textit{completion moment} prediction.
A similar approach was attempted in~\cite{Heidari2018}, however with full supervision and without temporal attention, where all frames contributed equally to the \textit{completion moment} detection.
We show that our proposed approach outperforms~\cite{Heidari2018} when fully supervised, but importantly is also able to detect \textit{completion} using weak video-level supervision. 

We evaluate our approach on selected actions from HMDB~\cite{HMDB}, UCF101~\cite{UCF101} and RGBD-AC~\cite{Heidari2016}. We show that learning temporal attention decreases the completion detection error, i.e. the relative distance between the predicted and ground truth \textit{completion moment}, by 15\% of the sequence length with weak supervision and by 3\% when fully supervised.

The remainder of this paper is organised as follows: related work in Sec.~\ref{sec:rel_work}, proposed method in Sec.~\ref{sec:ac_model}, experiments and results in Sec.~\ref{sec:res} and conclusion and future work in Sec.~\ref{sec:Conc}.

\section{Related Work}
\label{sec:rel_work}

In this section, we differentiate our work from approaches that attempted moment detection in actions, including for action completion. We also review works that utilised temporal attention learning, including for action localisation.

\noindent \textbf{Moment Detection:}
Temporal action detection from untrimmed videos~\cite{xu2017r,shou2016temporal,chao2018rethinking} involves localising start and end points of actions.
These works assume all actions are successfully completed, and do not consider incomplete attempts.
A few methods~\cite{becattini2017am,hoai2014max,dwibedi2019temporal,yeung2016glimpses,ma2016learning}, on the other hand, have adopted approaches that model action progression or detect particular key moments within actions.
Ma et al.~\cite{ma2016learning} detect an action by learning its progression through time. They devise a loss which maximises the margin between the correct action class and other classes as the action progresses further.
Hoai and De la Torre~\cite{hoai2014max} also detect actions in untrimmed sequences, where the action progression is modelled by a score function, learned using a Support Vector Machine classifier that peaks when the action ends. Similarly, Becattini~et~al.~\cite{becattini2017am} attempt to recognise actions by modelling their evolution through time, where the progress is assumed to be linear, reaching the highest at the end of the sequence.
Dwibedi et al.~\cite{dwibedi2019temporal} propose a self-supervised approach to learn temporal alignment between sequences based on the similarity between their frames and then observe an action's progression between key frames given the learnt alignment.
Yeung et al.~\cite{yeung2016glimpses} detect actions by looking at individual frames through the sequence, where the location of the next input frame is predicted relative to the current frame.
Although these works present a fine-grained analysis from the action progression, they also consider complete attempts and do not detect or localise the \textit{completion moment}.

\noindent \textbf{Action completion}~\cite{Heidari2016} differs from these works, as it focuses on the action's \textit{goal}. In~\cite{Heidari2018}, \textit{completion moment} detection was addressed using a classification-regression network which outputs frame-level predictions. These predictions are accumulated, using voting, to detect the \textit{completion moment}.
However, the method is fully supervised, requiring the \textit{completion moment} annotations for training. In contrast, we solve the same problem using only sequence-level \textit{complete} and \textit{incomplete} labels (i.e. \textit{weak labels}), through utilising temporal attention learning. 

\noindent \textbf{Attention Learning} has proven beneficial for research problems, such as image captioning~\cite{xu2015show,yao2015describing,chen2017sca}, object detection and tracking~\cite{caicedo2015semantic,denil2012learning,ba2014multiple} and person re-identification~\cite{haque2016recurrent,li2018diversity,xu2018attention}. Recently action recognition and localisation have also used attention networks to learn which spatial and/or temporal regions contain the most discriminative information. While some works~\cite{sharma2015action,girdhar2017attentional,li2018videolstm,wang2016hierarchical} have only focused on frame-level attention (spatial and motion), many others~\cite{du2018recurrent,song2017end,li2019unified,pei2017temporal,yeung2018every,wang2017untrimmednets,paul2018w,nguyen2018weakly} have also incorporated temporal attention in their models. They learn attention scores on the temporal dimension which are then used to weight the frames according to their importance to the final prediction. Of these, Pei et al.~\cite{pei2017temporal} introduce a recurrent unit for sequence classification in which a high attention score at each time step pushes the network to focus on the current observations rather than the past ones.  Song et al~\cite{song2017end} use LSTM for learning temporal attention from skeleton data in action recognition.
Du et al.~\cite{du2018recurrent} also propose an approach for action recognition using an LSTM with temporal softmax normalisation. 
Weighted observations, through learnt attention, from all frames in the sequence are combined to recognise the current frame's ongoing action.

For action localisation with weak supervision, several approaches have also attempted learning temporal attention, such as ~\cite{wang2017untrimmednets,paul2018w,nguyen2018weakly,li2019unified,yeung2018every}. For example, Yeung et~al.~\cite{yeung2018every} learn temporal attention for dense labelling in action localisation. Since they use trimmed sequences for training, but apply the learnt attention to localise actions in untrimmed sequences, their detection is considered weakly supervised. Other works, however, use untri\-mmed sequences in training.
Li et al. \cite{li2019unified} apply attention for action recognition and action detection
in untrimmed sequences, using features from multiple modalities as the input to the temporal attention LSTM before softmax normalisation.
Nguyen et al.~\cite{nguyen2018weakly} learn attention for action classification. They normalise the attention scores by a sigmoid function, and then use these to estimate the discriminative class-specific temporal regions for localising actions. Wang et al.~\cite{wang2017untrimmednets} predict the action's temporal extents by combining hard and soft selection methods, where the soft selection relies on the attention weights for the clip proposals sampled from the untrimmed sequences. In \cite{paul2018w}, the attention scores are first predicted as a temporal softmax on the class-wise activations and used during training. They then  apply a threshold on class-wise activations for localising actions.

In our method, we also use an LSTM for learning attention and a temporal softmax for its normalisation. However, our method  differs not only in the problem of \textit{completion moment} detection, but in how we accumulate evidence from all frames in the sequence based on learnt attention. We localise the \textit{completion moment} within trimmed sequences for both training and evaluation.

\section{Temporal Attention for \textit{Completion Moment} Detection}
\label{sec:ac_model}

\begin{figure*}[h]
\begin{center}
\includegraphics[width=.9\textwidth]{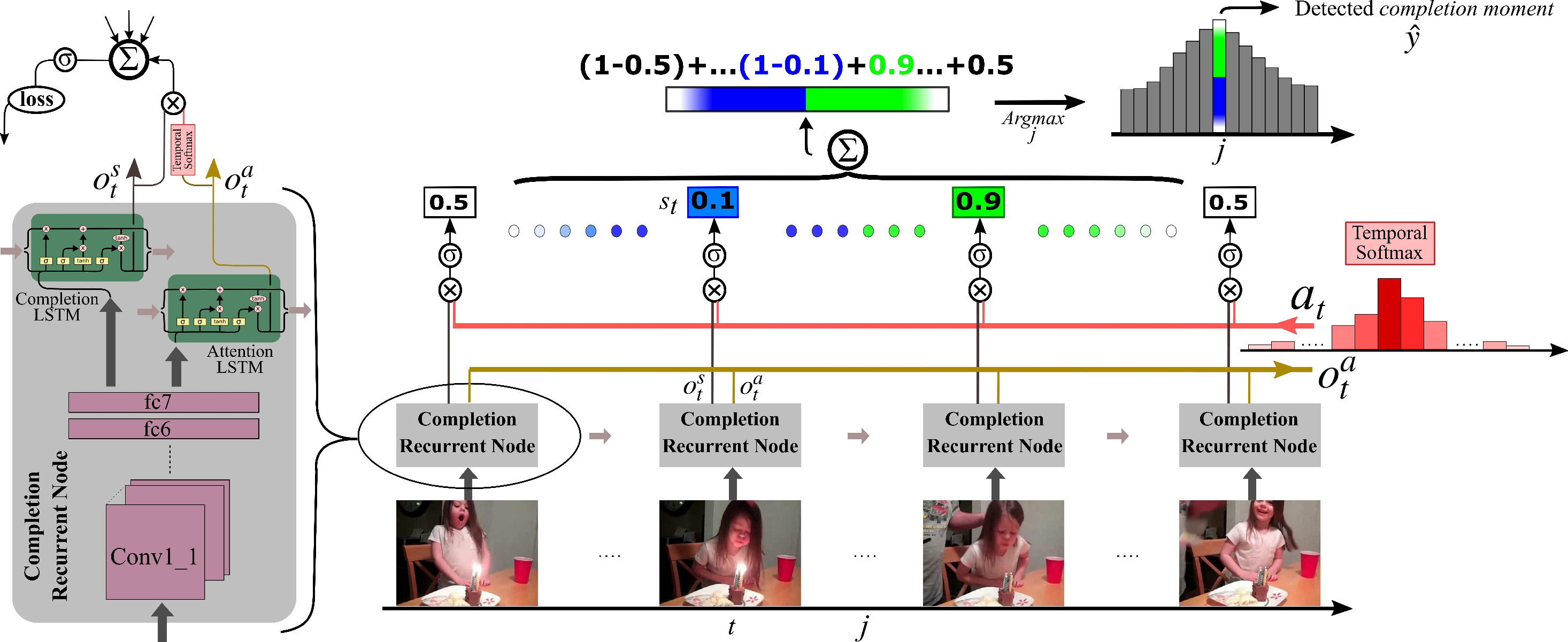}
\caption{Weakly supervised, the learnt attention scores (red distribution) and completion confidence scores are combined to infer the \textit{completion moment} $\hat{y}$ (Eq~\ref{eq:accum}). The frames depict a \textit{complete} sequence of action \textit{blowing candles} and colors green and blue represent the observed evidence for completion and incompletion, respectively.}
\label{fig:model_WS}
\end{center}
\end{figure*}

We now present our approach to \textbf{weakly-supervised} \textit{completion moment} detection, when video-level annotations only are  present.
Assume $X_i = \{x_i^1, \cdots x_i^T\}$ are the frames in a sequence of length $T$ where an action has been attempted, and $y_i \in \{0, 1\}$ is the binary video-level label, indicating whether the attempt has been successfully completed or not. 
Our method takes as input both complete and incomplete sequences of the same action.

To predict the \textit{completion moment} with weak supervision, we propose a network architecture that contains a convolutional frame-level feature extracting network, followed by two recurrent cells for completion prediction and temporal attention prediction, trained jointly with a cross-entropy loss function. Fig~\ref{fig:model_WS} depicts our architecture, showing the per-frame feature extraction and recurrent nodes (left) along with the training loss (top left). The frame-level predictions are then accumulated~(right) to infer the \textit{completion moment}.

For feature extraction, we train a convolutional network, by propagating the video-level label $y_i$ to \textit{all} frames in a video,
and optimise it using the cross-entropy loss, 
\begin{equation}
    L = \sum_{i = 1}^{M} \sum_{t = 1}^{T_i} - \big (y_i log f^c(f(x_i^t)) + (1-y_i) log (1-f^c(f(x_i^t))) \big)   ~,
\end{equation}
where $M$ is the number of sequences. The loss is optimised over all frames in all sequences, comparing the video-level labels $y_i$ against classification outputs $f^c()$, while frame-level features $f(x_i^t)$ are accordingly trained.
These learnt features form a good base for \textit{completion moment} detection, to be refined by the recurrent cells.
This is based on the realistic assumption that, up to the completion moment, both \textit{complete} and \textit{incomplete} sequences are indistinguishable.
However, after completion, there are appearance distinctions between the frames, to signify completion.

We train two recurrent models, namely LSTMs, jointly, one for temporal attention,~i.e.~to learn the relevance of each frame $t$ to \textit{completion moment} detection, $a_t$, and 
one to predict temporally-evidenced completion scores, $s_t$.
The temporal attention network is a standard LSTM, taking the features $f(x_t)$ as input - note that we simplified the notation $x_i^t$ to $x_t$ as the LSTM is trained and evaluated on one sequence. We compute the attention scores by applying a softmax function to the output nodes of this LSTM, $o^a_t$, across the temporal dimension, such that
\begin{equation}
    a_t = \frac{e^{o^a_t}}{\sum\limits_{j=1}^T e^{o^a_j}} ~ .
\end{equation}
The second LSTM, also takes the same input $f(x_t)$, and its output $o^s_t$ is then combined with the attention $a_t$ to produce completion scores per frame
\begin{equation}
    s_t = \frac{1}{1+e^{-a_t o^s_t}} ~ .
\end{equation}
The scores $s_t$ are the confidence of observing completion at frame $t$. In other words, a frame with a high $s_t$ has observed distinctive signatures for completion, making it more confident that the sequence has been \textit{completed}, with $1-s_t$ reflecting the confidence for incompletion.
We use these frame-level predictions to compute the \textit{completion moment}, such that
\begin{equation}
    \hat{y} = \arg\max_j \big (\sum_{t=1}^j (1-s_t) + \sum_{t=j+1}^T s_t \big )   ~ .
    \label{eq:accum}
\end{equation}
The predicted \textit{completion moment} $\hat{y}$ is one where the score for completion beyond  frame~$j$ as well as the score for incompletion before frame $j$ are the maximum.
  
During training, only video-level labels are available, and the ground-truth \textit{completion moment} is unknown. We thus train for sequence level prediction, such that 
\begin{eqnarray}
&\hat{y}_i^{tr} &=  \frac{1}{1+e^{-\sum_t (a_t o^s_t)}} ~ , \\
&L &=\sum_i -(y_i \hat{y}_i^{tr}+ (1-y_i) (1-\hat{y}_i^{tr})) ~ ,
\end{eqnarray}
where $y_i^{tr}$ indicates whether the sequence has been completed, \textit{somewhere} along its frames. These predictions are optimised against the video-level completion labels, for all sequences. Note that, using $a_t$ in the training loss makes the model learn to weight highly the temporal regions which contain discriminative evidence for completion.

\begin{figure*}[h]
\begin{center}
\includegraphics[width=.9\textwidth]{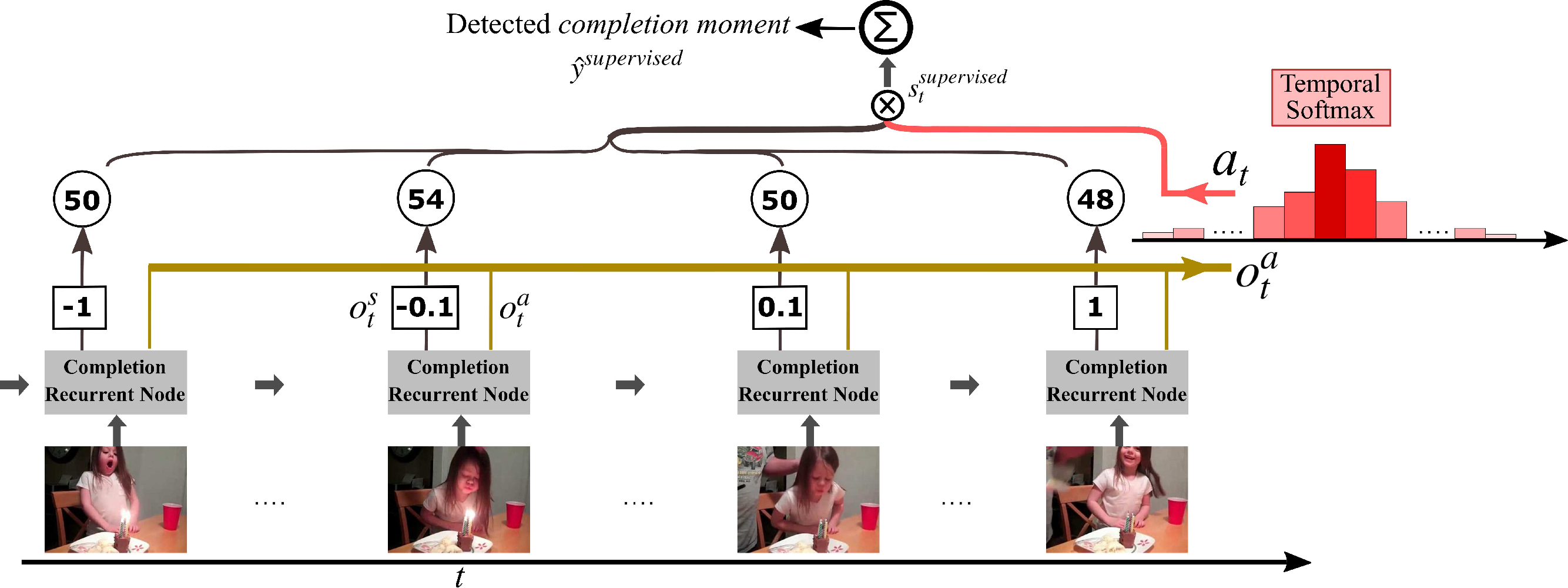}
\caption{In the Supervised model, the sequence level \textit{completion moment} is the weighted average of the frame-level predictions, using the attention scores.}

\label{fig:model_S}
\end{center}
\end{figure*}

While focusing on weakly-supervised \textit{completion moment} detection, we also evaluate our proposed architecture in a \textbf{supervised} approach. We similarly combine completion detection with temporal attention, when supervision for the \textit{completion moment} is available.
We thus train the output of the confidence scores $o^s_t$ in the same way as the regression-based supervision in~\cite{Heidari2018}, using the relative distance $r_t = \frac{t-\tau_i}{\tau_i}$ between the frame $t$ and the ground-truth \textit{completion moment} $\tau_i$, allowing the approaches to be directly comparable. The sequence-level loss $L_i$ would then be:
\begin{equation}
L_i^{supervised} = \sum_{t=1}^{T_i}  a_t  (o^s_t-r_t)^2 .
\label{eq:L_s_att}
\end{equation}
Using these scores,
\begin{equation}
    s_t^{supervised} = a_t \frac{t}{o_t^s+1},
\end{equation}
estimates the \textit{completion moment} from each frame, weighted by the learnt attention scores. The sequence-level \textit{completion moment} is finally predicted as
\begin{equation}
    \hat{y}^{supervised} = \sum_{t=1}^{T} s_t.
\end{equation}
Fig.~\ref{fig:model_S} illustrates the supervised completion detection where the frame-level evidences are accumulated across the sequence during inference.

\section{Experimental Results}
\label{sec:res}
\textbf{Dataset and Implementation Details -- }
We evaluate our approach on the 16 actions used in~\cite{Heidari2018} as the only prior work to attempt \textit{completion moment} detection, and using the publicly available annotations provided by~\cite{Heidari2018}. These actions have been collected from three public datasets: HMDB~\cite{HMDB}, UCF101~\cite{UCF101} and RGBD-AC~\cite{Heidari2016}. As stated in~\cite{Heidari2018}, these actions cover sport-based and daily actions, for which \textit{completion} can be defined, and include both complete and incomplete sequences for training.
We report results on all 16 actions when supervised. However, in the weakly supervised setting, we require sufficient \textit{incomplete} sequences per action to be able to train with only video-level weak labels. Of these 16 actions, we only evaluate on 10 actions which have both \textit{complete} and \textit{incomplete} sequences, while the remaining 6 have less than 5\% incomplete sequences.

For feature extraction, we used the spatial stream of VGG-16 architecture, pre-trained on UCF101. We then fine-tuned it for 20 epochs to acquire frame-level features. The learning rate was started at $10^{-3}$, divided by 10 at epochs 3 and 5.
The features were extracted from the output of the $fc7$ layer. Both LSTM cells (attention and \textit{completion moment} prediction) had a single layer with 128 hidden units.
When fully supervised, we first trained the completion prediction LSTM $s_t$ for 10 epochs for stability, then jointly trained both LSTMs for 5 more epochs.
When weakly-supervised, we initialised both LSTMs from random and trained them jointly for 10 epochs. The learning rates for the LSTM training in both approaches was $10^{-2}$ for the first 5 epochs and then was divided by 10 for the rest.
For temporal prediction, we normalised the sequences to a fixed length, equal to the minimum length of any sequence in that action. 
Note that our method is not dependent on the sequence length and thus is robust to any other pre-specified length. Additionally, the attention scores were normalised between zero and one and those less than 0.5 were truncated to 0 during inference .

\noindent \textbf{Evaluation Metrics --} As in~\cite{Heidari2018}, we report the accuracy as the average percentage of frames that are correctly labeled into \textit{pre-} and \textit{post-completion}, given the ground-truth $\tau_i$ and the predicted \textit{completion moment} $\hat{y}_i$, such that
\begin{equation}
\begin{split}
\text{Accuracy} = \frac{1}{M} \sum_{i=1}^M \ \frac{1}{T_i} \sum_{t=1}^{T_i} \ \big[ & (t < \hat{y}_i \ \land \ t < \tau_i) \\ \lor \ 
& (t \ge \hat{y}_i \ \land \ t \ge \tau_i)\big] ~ .
\end{split}
\end{equation}
We also report $RD$ as the relative distance between the predicted and ground truth \textit{completion moment}, averaged on all sequences.

\begin{table}[t]
\centering
\resizebox{\linewidth}{!}{
\begin{tabular}{|c|c|c|c|c||c|c|}
\cline{3-7}
\multicolumn{2}{c|}{ } & \bf Incomplete & \multicolumn{2}{|c||}{\bf Accuracy} & \multicolumn{2}{c|}{$RD$}\\ \cline{4-7}
\multicolumn{2}{c|}{ } & \bf \% & \textbf{WS-U} & \textbf{WS-Att}& \textbf{WS-U} & \textbf{WS-Att} \\ \hline
HMDB & \textit{pick} & 22.4 & 34.8 & \textbf{48.6} & 0.65 &  \textbf{0.51} \\ 
\hline \hline
\multirow{3}{*}{UCF101} & \textit{basketball} & 23.9 & 38.2 & \textbf{58.0} & 0.62 &  \textbf{0.42} \\ \cline{2-7}
& \textit{soccer penalty} & 30.7 & 34.4 & \textbf{55.6} & 0.66 & \textbf{0.44} \\ \cline{2-7}
& \textit{blowing candles} & 45.9 & 43.8 & \textbf{70.9} & 0.56 &  \textbf{0.29} \\ 
\hline \hline
\multirow{6}{*}{RGBD-AC} & \textit{switch} & 47.8 & 86.9 & \textbf{89.9} & 0.13 &  \textbf{0.1} \\ \cline{2-7}
& \textit{plug} & 49.3 & 62.0 & \textbf{78.6} & 0.38 &  \textbf{0.21} \\ \cline{2-7}
& \textit{open} & 47.1 & 74.6 & \textbf{77.1} & 0.25 & \textbf{0.23} \\ \cline{2-7}
& \textit{pull} & 52.1 & 79.1 & \textbf{83.8} & 0.21 & \textbf{0.16} \\ \cline{2-7}
& \textit{pick} & 52.2 & 58.4 & \textbf{83.0} & 0.42 & \textbf{0.17} \\ \cline{2-7}
& \textit{drink} & 48.5 & 57.2 & \textbf{69.6}  & 0.43 & \textbf{0.30} \\
\hline \hline
\multicolumn{2}{|c|}{\textbf{total}} & 39.3 & 53.8 & \textbf{69.4} & 0.46 & \textbf{0.31}\\ \hline
\end{tabular}}
\caption{Results comparing \textbf{weakly supervised} \textit{completion moment} detection with and without temporal attention learning.}
\label{tab:res_WS}
\end{table}

\begin{table}[t]
\centering
\resizebox{\linewidth}{!}{
\begin{tabular}{|c|c|c|c|c||c|c|c|}
\cline{3-8}
\multicolumn{2}{c|}{ } & \multicolumn{3}{c||}{\bf Accuracy} & \multicolumn{3}{c|}{$RD$}\\ \cline{3-8}
\multicolumn{2}{c|}{ } & \textbf{\cite{Heidari2018}} & \textbf{S-U} & \textbf{S-Att} & \textbf{\cite{Heidari2018}} & \textbf{S-U} & \textbf{S-Att}\\ \hline
\multirow{6}{*}{\rotatebox[origin=c]{90}{HMDB}} & \textit{catch} & 80.5 & 82.3 & \textbf{83.5} & 0.20 & 0.18 & \textbf{0.17}\\ \cline{2-8}
&\textit{drink} & 78.0 & 80.3 & \textbf{81.1}  & 0.22 & 0.20 & \textbf{0.19} \\ \cline{2-8}
&\textit{pick}  & 79.9 & 81.8 & \textbf{83.2} & 0.20 & 0.18 & \textbf{0.17}\\ \cline{2-8}
&\textit{pour} & \textbf{80.0} & 77.8 & 78.3 & \textbf{0.20} & 0.22 & 0.22\\ \cline{2-8}
&\textit{throw} & 74.6 & 76.9 & \textbf{78.3} & 0.25 & 0.23 & \textbf{0.22}\\ %
\hline \hline

\multirow{6}{*}{\rotatebox[origin=c]{90}{UCF101}} & \textit{basketball} & 79.5 & 82.8 & \textbf{83.6} & 0.20 & 0.17 & \textbf{0.16} \\ \cline{2-8}
& \textit{blowing candles} & 84.2 & 89.9 & \textbf{90.1} & 0.16 & \textbf{0.10} & \textbf{0.10}\\ \cline{2-8}
& \textit{frisbee catch} & 78.3 & 86.6 & \textbf{86.9} & 0.22 & \textbf{0.13} & \textbf{0.13} \\ \cline{2-8}
& \textit{pole vault} & 88.4 & 88.4 & \textbf{89.7} & 0.12 & 0.12 & \textbf{0.10} \\ \cline{2-8}
& \textit{soccer penalty} & 87.1 & 87.0 & \textbf{87.9} & 0.13 & 0.13 & \textbf{0.12} \\ %
\hline \hline

\multirow{6}{*}{\rotatebox[origin=c]{90}{RGBD-AC}} & \textit{switch} & 98.1 & 94.6 & \textbf{98.2} & \textbf{0.02} & 0.05 & \textbf{0.02} \\ \cline{2-8}
& \textit{plug} & 96.1 & 95.1 & \textbf{96.8} & 0.04 & 0.05 & \textbf{0.03} \\ \cline{2-8}
& \textit{open} & 86.7 & 88.2 & \textbf{91.3} & 0.13 & 0.12 & \textbf{0.09} \\ \cline{2-8}
& \textit{pull} & 94.1 & 92.4 & \textbf{95.1} & 0.06 & 0.08 & \textbf{0.05} \\ \cline{2-8}
& \textit{pick} & \textbf{93.2} & 90.5 & 92.0 & \textbf{0.07} & 0.09 & 0.08 \\ \cline{2-8}
& \textit{drink} & 90.9 & 89.0 & \textbf{91.2} & \textbf{0.09} & 0.11 & \textbf{0.09} \\
\hline \hline
\multicolumn{2}{|c|}{\textbf{total}} & 84.9 & 86.1 & \textbf{87.4} & 0.15 & 0.14 & \textbf{0.12}\\ \hline

\end{tabular}}
\caption{Results comparing \textbf{fully supervised} \textit{completion moment} detection with and without temporal attention learning.}
\label{tab:res_S}
\end{table}

\begin{equation}\label{eq:rel_dist}
RD = \frac{1}{M}\sum\limits_{i = 1}^M{\frac{||\hat{y}_i-\tau_i||}{T_i}} \ .
\end{equation}

\begin{figure*}[t]
\begin{center}
\includegraphics[width=.95\textwidth]{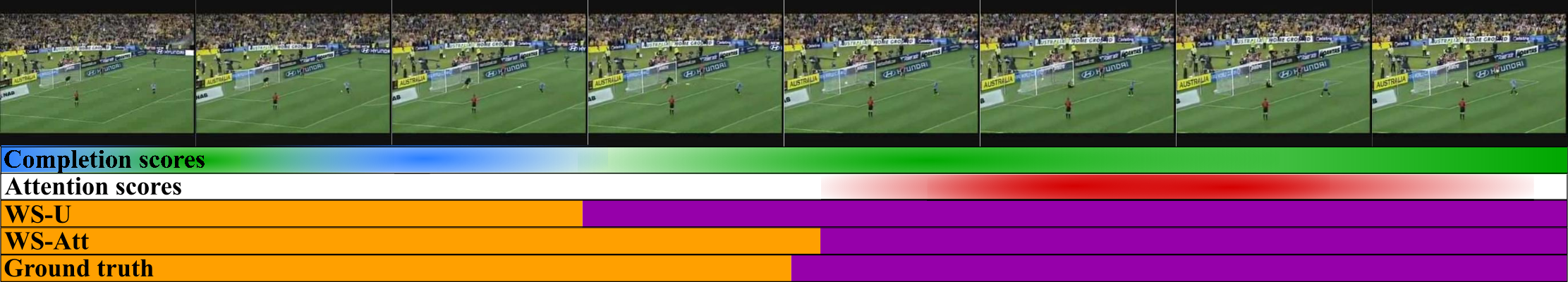}\\
\includegraphics[width=.95\textwidth]{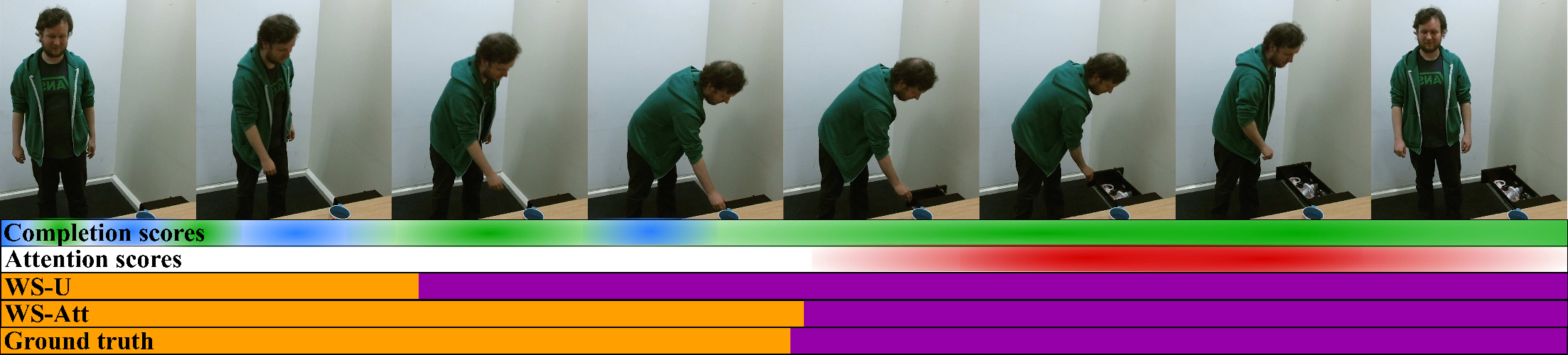}\\
\includegraphics[width=.95\textwidth]{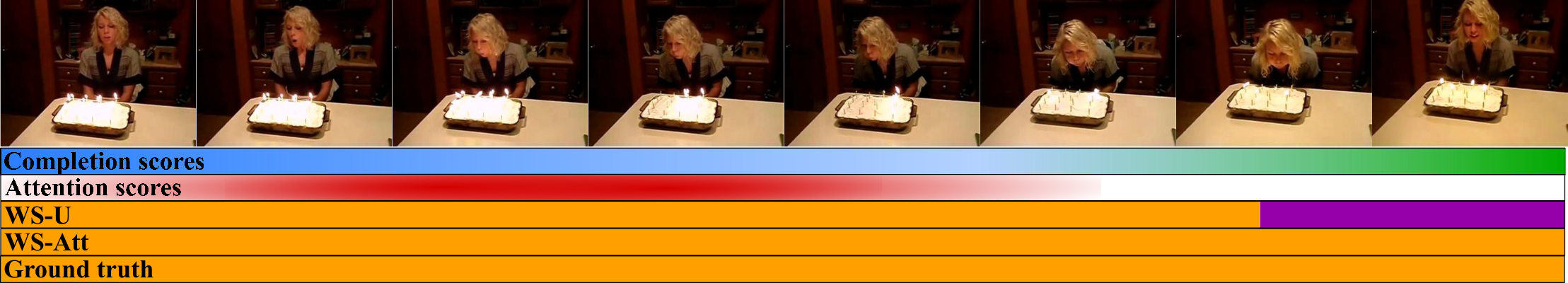}\\ \vspace{-3pt}
Examples of success\\
\vspace{10pt}
\includegraphics[width=.95\textwidth]{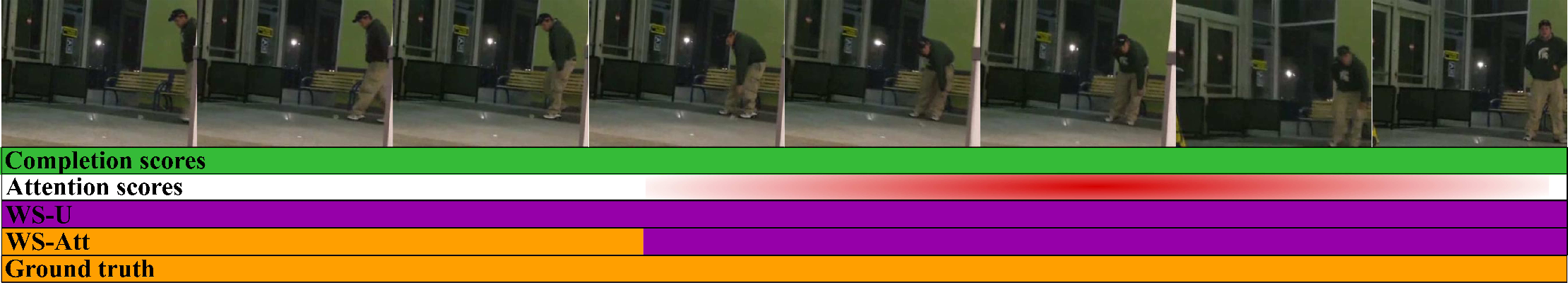}\\ \vspace{-3pt}
Example of failure
\caption{Qualitative results for weakly supervised \textit{completion moment} detection. Top to bottom: UCF101-\textit{soccer penalty}, RGBD-AC-\textit{pull}, UCF101-\textit{blowing candles} and HMDB-\textit{pick}, respectively.}
\label{fig:res_WS}
\end{center}
\end{figure*}

\begin{figure*}[t]
\begin{center}
\includegraphics[width=.9\linewidth]{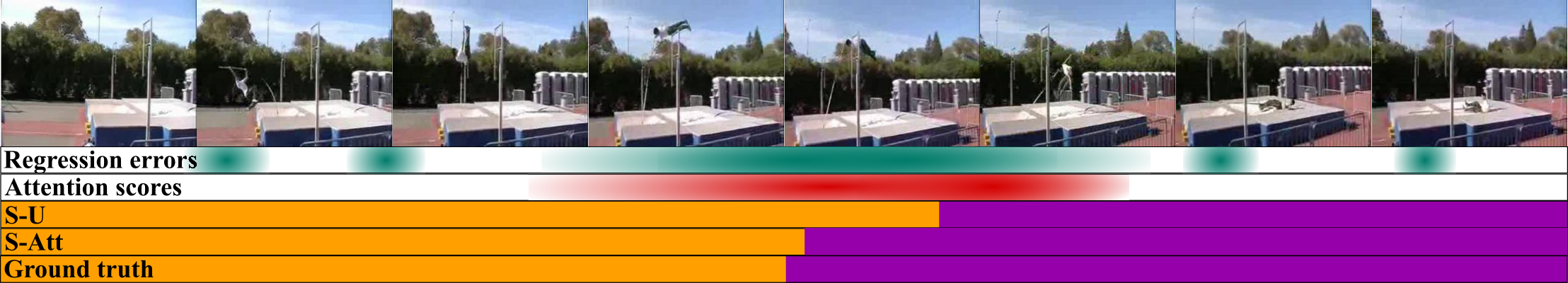}\\
\includegraphics[width=.9\linewidth]{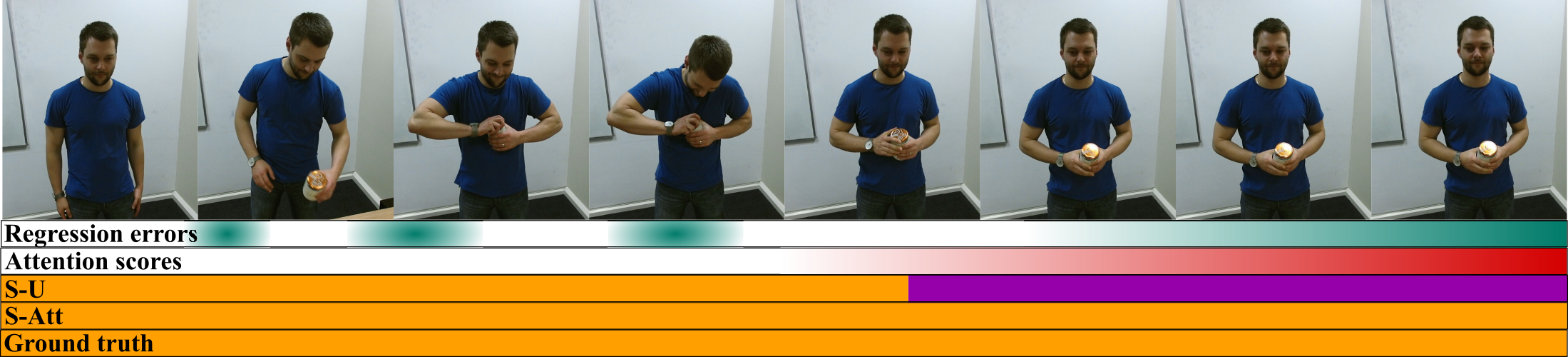}\\ \vspace{-3pt}
Examples of success\\
\vspace{10pt}
\includegraphics[width=.9\linewidth]{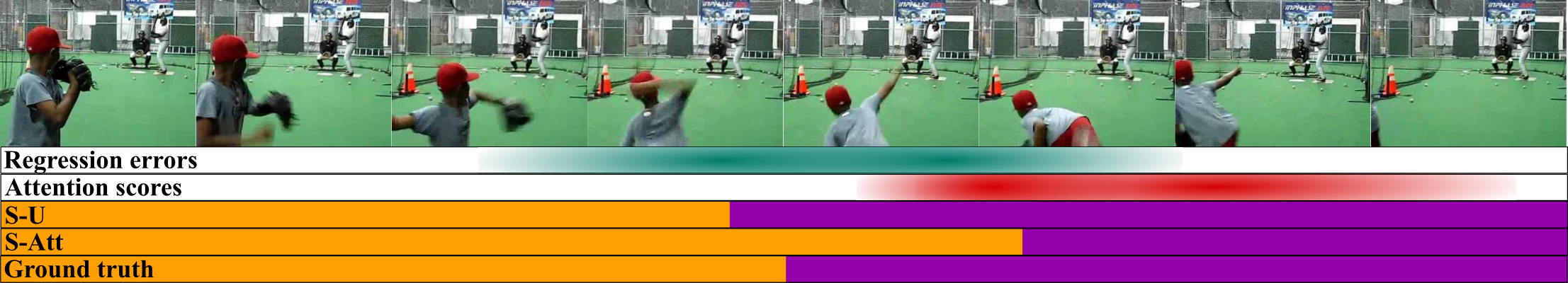}\\
\includegraphics[width=.9\linewidth]{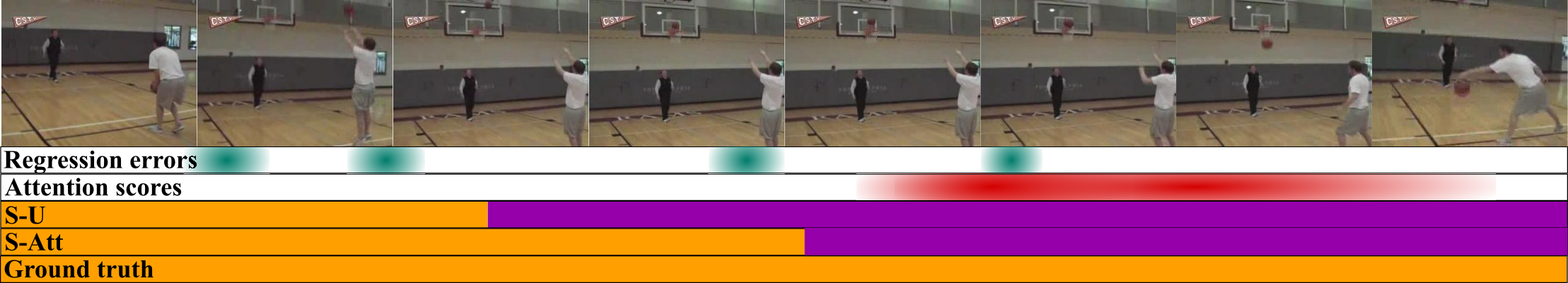}\\ \vspace{-3pt}
Examples of failalure\\
\caption{Qualitative results, using supervised learning. Top left to bottom right: UCF101-\textit{pole vault}, RGBD-AC-\textit{open}, HMDB-\textit{throw} and UCF101-\textit{basketball}, respectively.}
\label{fig:res_S}
\end{center}
\end{figure*}
\begin{figure*}[t]
\begin{center}
\begin{tabular}{ccc}
\includegraphics[width=.32\textwidth]{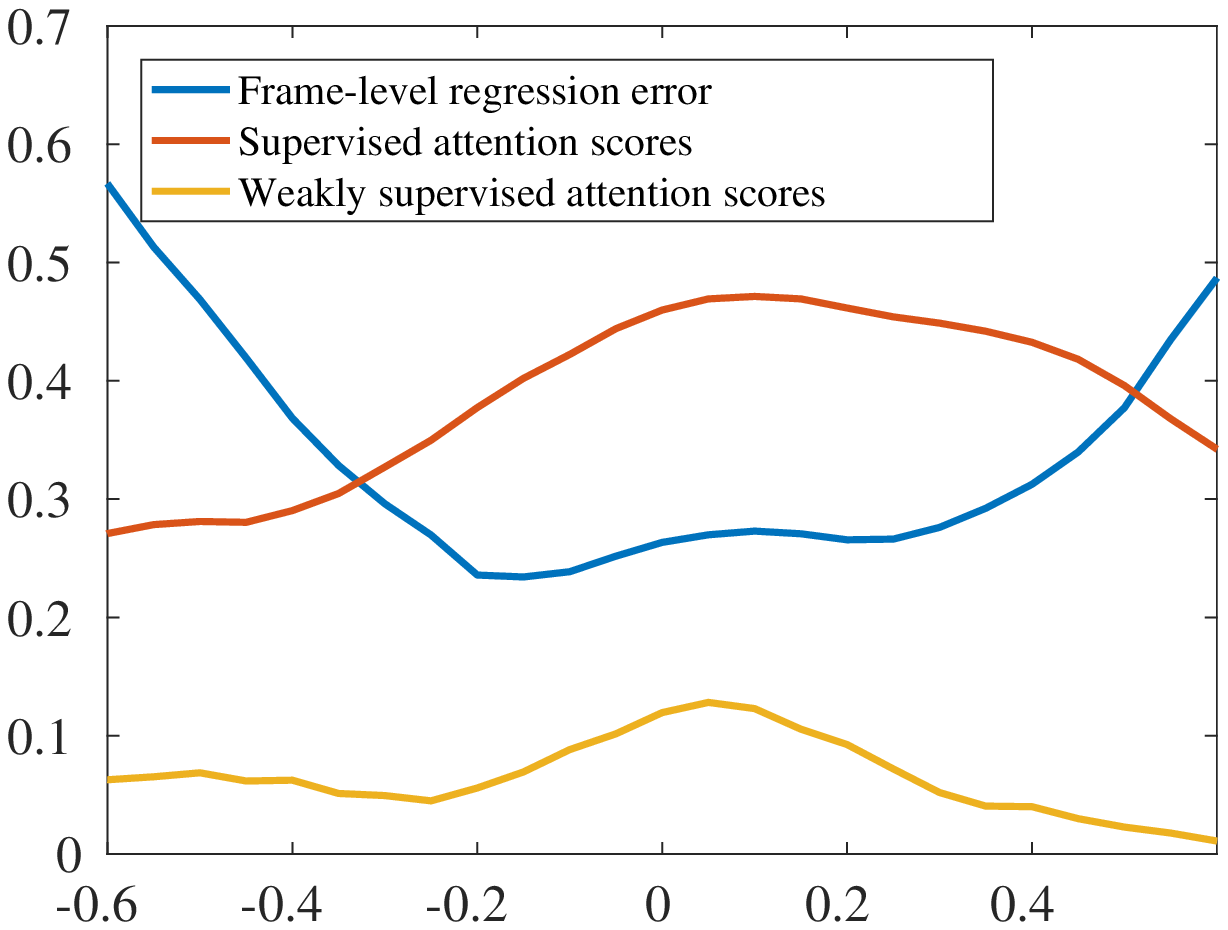}
&\includegraphics[width=.32\textwidth]{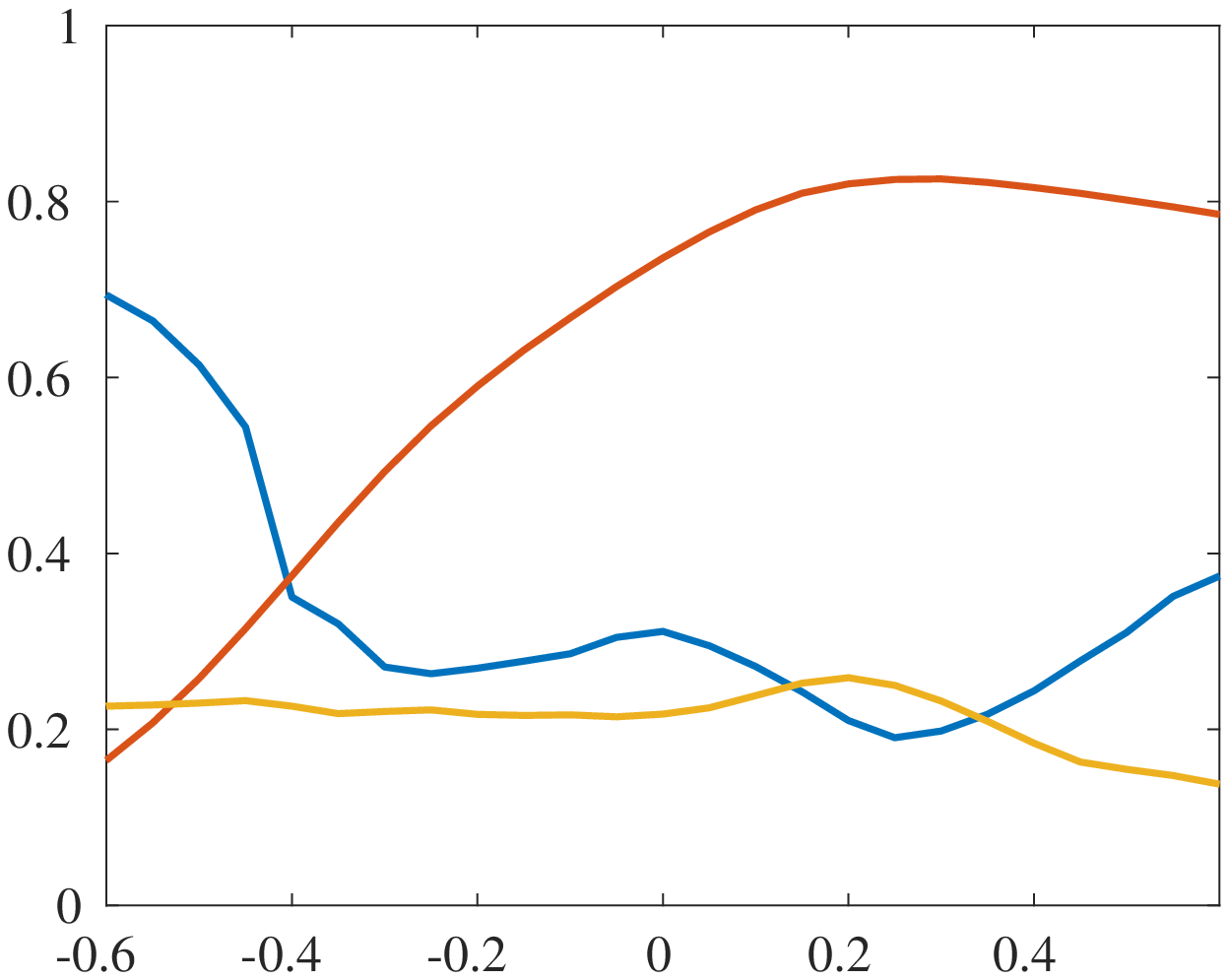}
&\includegraphics[width=.32\textwidth]{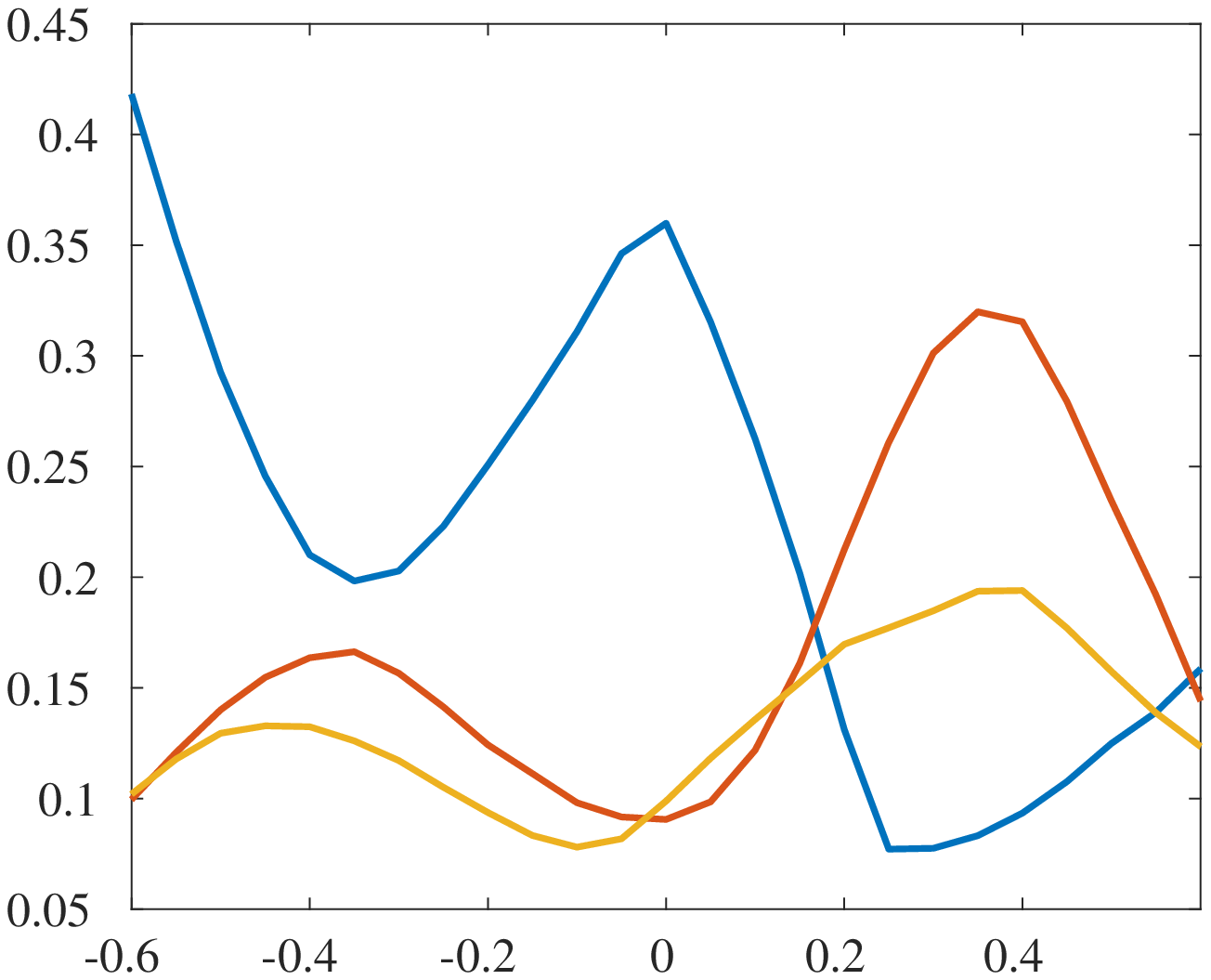}\\
HMDB & UCF101 & RGBD-AC
\end{tabular}
\caption{Frame-level regression errors (in blue) are depicted against the attention scores in both fully (red) and weakly (orange) supervised approaches. Note that `0' on the x-axis indicates the \textit{completion moment} for all \textit{complete} sequences, and the end of the sequence for all \textit{incomplete} sequences.}

\label{fig:RD_online}
\end{center}
\end{figure*}

\noindent \textbf{Weakly Supervised Completion Detection -- }
Table~\ref{tab:res_WS} shows the results of our proposed method for weakly supervised \textit{completion moment} detection using uniform attention (WS-U) as well as with learnt temporal attention (WS-Att). In WS-U, we do not learn attention, and use uniform weighting in inference. Learning temporal attention improves results for all actions and both metrics.
For actions with a smaller percentage of \textit{incomplete} sequences, i.e.~HMDB-\textit{pick}, UCF101-\textit{basketball} and UCF101-\textit{soccer penalty}, the performance is lower for both metrics, though temporal attention consistently improves the results. {In total, i.e. on all sequences from the three datasets, RD drops to 0.31 with WS-Att.}

We also present some qualitative results for the weakly supervised approach in Fig.~\ref{fig:res_WS} where the first bar depicts the completion scores $s_t$ - with green and blue representing the observed evidence for completion and incompletion respectively. The attention is shown in red, and results in orange and purple represent \textit{pre} and \textit{post-completion} labels, respectively.
In the first two sequences, the temporal attention significantly improves the results by correctly weighting the frames after completion where discriminative featuers are observed.
In the third example from action \textit{blowing candles}, while WS-U has been misled by the completion scores at the end of the sequence, WS-Att correctly detected no completion.
The last sequence shows a failure case for action \textit{pick}. We believe this would be improved with more \textit{incomplete} sequences during training.

\noindent \textbf{Supervised Temporal Attention Learning -- }
Table~\ref{tab:res_S} shows the results of the supervised approach, compared to the R-R method in \cite{Heidari2018}, which is comparable to ours as it does not use frame-level \textit{pre/post-completion} classification, but directly predicts the \textit{completion moment}. We also compare uniform weighting (S-U) to the learnt attention~(S-Att).
Learning temporal attention outperforms uniform weighting on all 16 actions, and outperforms the baseline on 14 out of the 16 actions. {In total, RD drops to 0.12 with S-Att.}

We present qualitative results for our method, when supervised, in Fig~\ref{fig:res_S}.
The first bar represents the frame-level regression \textbf{error}, i.e. {$||o^s_t-r_t||$} (darker is lower error).
The examples of success (left) show two sequences from actions UCF101-\textit{pole vault} and RGBD-AC-\textit{open}.
Temporal attention improves the \textit{completion moment} detection for both complete (top left) and incomplete (bottom left) sequences, as high attention correctly aligns to regions with small prediction error.
The examples of failure (right) represent two sequences from actions HMDB-\textit{throw} and UCF101-\textit{basketball}, where attention has not been able to pick the regions with small error. 
In the \textit{basketball} example, the sequence is detected as \textit{complete} with and without attention, despite being \textit{incomplete}.

\noindent \textbf{Frame-level Analysis -- }
We plot the frame-level errors, as well as the attention scores, averaged for all actions, on the three datasets in Fig.~\ref{fig:RD_online}.
The figure shows lower prediction errors (blue), both \textit{before} and \textit{after} the completion moment, in two clear minimas. Increased confusion around the completion moment comes from the very similar features before \textit{completion moment}. We also show the learnt temporal attention for both supervised (red) and weakly supervised  (orange) approaches. Generally, higher attention corresponds to lower prediction error - signifying that these frames will have a higher impact in the overall \textit{completion moment} prediction. When weakly-supervised, the attention scores are comparable to full-supervision though understandably softer attention is learnt.

\section{Conclusion and Future Work}
\label{sec:Conc}

In this paper, we proposed a method to detect the \textit{completion moment} in a variety of actions, suitable for both weakly-supervised and fully supervised sequences. In weak-supervision, video-level labels of \textit{completion} or \textit{incompletion} are only required, for the same action. When a sufficient number of incomplete sequences is available during training, our approach, (1) learns discriminative features for frames pre- and post- completion, by propagating video-level labels to individual frames, (2) learns temporal attention, to weight discriminative frame-level features, and then (3) accumulates evidence for completion, weighted by the learnt attention, from all frames to predict the \textit{completion moment}, or identify the attempt as \textit{incomplete}.
We evaluated our approach on 16 actions (with full supervision) and 10 actions (with weak supervision), from 3 datasets. When weakly-supervised, learning attention significantly improved the results on all tested actions. Under full supervision, we outperform prior work~\cite{Heidari2018} on 14 out of the 16 actions.

For future work, we aim to augment the temporal attention with within-frame spatial attention to learn image regions that are most discriminative for completion. We will also combine our soft attention with hard attention mechanisms, similar to~\cite{wang2017untrimmednets}. Further, we will investigate \textit{completion moment} detection from untrimmed sequences, which contain multiple action instances.

\section*{Acknowledgements}
\label{sec:Ackn}

The 1st author wishes to thank the University of Bristol for partial funding of her studies. Public datasets were used in this work.

{\small
\bibliographystyle{ieee}
\bibliography{egbib}

\begin{thebibliography}{10}\itemsep=-1pt

\bibitem{ba2014multiple}
J.~Ba, V.~Mnih, and K.~Kavukcuoglu.
\newblock Multiple object recognition with visual attention.
\newblock {\em arXiv preprint arXiv:1412.7755}, 2014.

\bibitem{becattini2017am}
F.~Becattini, T.~Uricchio, L.~Ballan, L.~Seidenari, and A.~Del~Bimbo.
\newblock Am {I} done? {P}redicting action progress in videos.
\newblock {\em arXiv preprint arXiv:1705.01781}, 2017.

\bibitem{caicedo2015semantic}
J.~Caicedo and S.~Lazebnik.
\newblock Semantic guidance of visual attention for localizing objects in
  scenes.
\newblock In {\em ICCV}, 2015.

\bibitem{carreira2017quo}
J.~Carreira and A.~Zisserman.
\newblock Quo {V}adis, {A}ction recognition? {A} new model and the kinetics
  dataset.
\newblock In {\em CVPR}, 2017.

\bibitem{chao2018rethinking}
Y.~Chao, S.~Vijayanarasimhan, B.~Seybold, D.~Ross, J.~Deng, and R.~Sukthankar.
\newblock Rethinking the faster {R-CNN} architecture for temporal action
  localization.
\newblock In {\em CVPR}, 2018.

\bibitem{chen2017sca}
L.~Chen, H.~Zhang, J.~Xiao, L.~Nie, J.~Shao, W.~Liu, and T.~Chua.
\newblock {SCA-CNN}: Spatial and channel-wise attention in convolutional
  networks for image captioning.
\newblock In {\em CVPR}, 2017.

\bibitem{denil2012learning}
M.~Denil, L.~Bazzani, H.~Larochelle, and N.~de~Freitas.
\newblock Learning where to attend with deep architectures for image tracking.
\newblock {\em Neural {C}omputation}, 2012.

\bibitem{du2018recurrent}
W.~Du, Y.~Wang, and Y.~Qiao.
\newblock Recurrent spatial-temporal attention network for action recognition
  in videos.
\newblock {\em IEEE Transactions on Image Processing}, 2018.

\bibitem{dwibedi2019temporal}
D.~Dwibedi, Y.~Aytar, J.~Tompson, P.~Sermanet, and A.~Zisserman.
\newblock Temporal cycle-consistency learning.
\newblock {\em arXiv preprint arXiv:1904.07846}, 2019.

\bibitem{girdhar2017attentional}
R.~Girdhar and D.~Ramanan.
\newblock Attentional pooling for action recognition.
\newblock In {\em NIPS}, 2017.

\bibitem{girdhar2017actionvlad}
R.~Girdhar, D.~Ramanan, A.~Gupta, J.~Sivic, and B.~Russell.
\newblock Action{VLAD}: Learning spatio-temporal aggregation for action
  classification.
\newblock In {\em CVPR}, 2017.

\bibitem{haque2016recurrent}
A.~Haque, A.~Alahi, and L.~Fei-Fei.
\newblock Recurrent attention models for depth-based person identification.
\newblock In {\em CVPR}, 2016.

\bibitem{Heidari2016}
F.~Heidarivincheh, M.~Mirmehdi, and D.~Damen.
\newblock Beyond action recognition: Action completion in {RGB-D} data.
\newblock In {\em BMVC}, 2016.

\bibitem{Heidari2018}
F.~Heidarivincheh, M.~Mirmehdi, and D.~Damen.
\newblock Action completion: a temporal model for moment detection.
\newblock {\em BMVC}, 2018.

\bibitem{hoai2014max}
M.~Hoai and F.~De~la Torre.
\newblock Max-margin early event detectors.
\newblock {\em International {J}ournal of {C}omputer {V}ision (IJCV)}, 2014.

\bibitem{karpathy2014large}
A.~Karpathy, G.~Toderici, S.~Shetty, T.~Leung, R.~Sukthankar, and L.~Fei-Fei.
\newblock Large-scale video classification with convolutional neural networks.
\newblock In {\em CVPR}, 2014.

\bibitem{HMDB}
H.~Kuehne, H.~Jhuang, E.~Garrote, T.~Poggio, and T.~Serre.
\newblock {HMDB}: {A} large video database for human motion recognition.
\newblock In {\em ICCV}, 2011.

\bibitem{laptev2005space}
I.~Laptev.
\newblock On space-time interest points.
\newblock {\em International {J}ournal of {C}omputer {V}ision (IJCV)}, 2005.

\bibitem{li2019unified}
D.~Li, T.~Yao, L.~Duan, T.~Mei, and Y.~Rui.
\newblock Unified spatio-temporal attention networks for action recognition in
  videos.
\newblock {\em IEEE Transactions on Multimedia}, 2019.

\bibitem{li2018diversity}
S.~Li, S.~Bak, P.~Carr, and X.~Wang.
\newblock Diversity regularized spatiotemporal attention for video-based person
  re-identification.
\newblock In {\em CVPR}, 2018.

\bibitem{li2018videolstm}
Z.~Li, K.~Gavrilyuk, E.~Gavves, M.~Jain, and C.~Snoek.
\newblock Video{LSTM} convolves, attends and flows for action recognition.
\newblock {\em Computer Vision and Image Understanding (CVIU)}, 2018.

\bibitem{ma2016learning}
S.~Ma, L.~Sigal, and S.~Sclaroff.
\newblock Learning activity progression in {LSTM}s for activity detection and
  early detection.
\newblock In {\em CVPR}, 2016.

\bibitem{moltisanti19action}
D.~Moltisanti, S.~Fidler, and D.~Damen.
\newblock {A}ction {R}ecognition from {S}ingle {T}imestamp {S}upervision in
  {U}ntrimmed {V}ideos.
\newblock In {\em Computer Vision and Pattern Recognition (CVPR)}, 2019.

\bibitem{moltisanti2017trespassing}
D.~Moltisanti, M.~Wray, W.~Mayol-Cuevas, and D.~Damen.
\newblock Trespassing the boundaries: Labeling temporal bounds for object
  interactions in egocentric video.
\newblock In {\em ICCV}, 2017.

\bibitem{nguyen2018weakly}
P.~Nguyen, T.~Liu, G.~Prasad, and B.~Han.
\newblock Weakly supervised action localization by sparse temporal pooling
  network.
\newblock In {\em CVPR}, 2018.

\bibitem{paul2018w}
S.~Paul, S.~Roy, and A.~Roy-Chowdhury.
\newblock W-{TALC}: {W}eakly-supervised temporal activity localization and
  classification.
\newblock In {\em ECCV}, 2018.

\bibitem{pei2017temporal}
W.~Pei, T.~Baltrusaitis, D.~Tax, and L.~Morency.
\newblock Temporal attention-gated model for robust sequence classification.
\newblock In {\em CVPR}, 2017.

\bibitem{sharma2015action}
S.~Sharma, R.~Kiros, and R.~Salakhutdinov.
\newblock Action recognition using visual attention.
\newblock {\em arXiv preprint arXiv:1511.04119}, 2015.

\bibitem{shou2016temporal}
Z.~Shou, D.~Wang, and S.~Chang.
\newblock Temporal action localization in untrimmed videos via multi-stage
  {CNN}s.
\newblock In {\em CVPR}, 2016.

\bibitem{sigurdsson2016much}
G.~Sigurdsson, O.~Russakovsky, A.~Farhadi, I.~Laptev, and A.~Gupta.
\newblock Much ado about time: Exhaustive annotation of temporal data.
\newblock {\em arXiv preprint arXiv:1607.07429}, 2016.

\bibitem{simonyan2014two}
K.~Simonyan and A.~Zisserman.
\newblock Two-stream convolutional networks for action recognition in videos.
\newblock In {\em NIPS}, 2014.

\bibitem{song2017end}
S.~Song, C.~Lan, J.~Xing, W.~Zeng, and J.~Liu.
\newblock An end-to-end spatio-temporal attention model for human action
  recognition from skeleton data.
\newblock In {\em AAAI}, 2017.

\bibitem{UCF101}
K.~Soomro, A.~Roshan~Zamir, and M.~Shah.
\newblock A dataset of 101 human actions classes from videos in the wild.
\newblock {\em arXiv preprint arXiv:1212.0402}, 2012.

\bibitem{wang2013action}
H.~Wang and C.~Schmid.
\newblock Action recognition with improved trajectories.
\newblock In {\em ICCV}, 2013.

\bibitem{wang2013motionlets}
L.~Wang, Y.~Qiao, and X.~Tang.
\newblock Motionlets: {M}id-level 3{D} parts for human motion recognition.
\newblock In {\em CVPR}, 2013.

\bibitem{wang2017untrimmednets}
L.~Wang, Y.~Xiong, D.~Lin, and L.~Van~Gool.
\newblock Untrimmednets for weakly supervised action recognition and detection.
\newblock In {\em CVPR}, 2017.

\bibitem{wang2016hierarchical}
Y.~Wang, S.~Wang, J.~Tang, N.~O'Hare, Y.~Chang, and B.~Li.
\newblock Hierarchical attention network for action recognition in videos.
\newblock {\em arXiv preprint arXiv:1607.06416}, 2016.

\bibitem{xu2017r}
H.~Xu, A.~Das, and K.~Saenko.
\newblock {R-C3D}: Region convolutional 3{D} network for temporal activity
  detection.
\newblock In {\em ICCV}, 2017.

\bibitem{xu2018attention}
J.~Xu, R.~Zhao, F.~Zhu, H.~Wang, and W.~Ouyang.
\newblock Attention-aware compositional network for person re-identification.
\newblock In {\em CVPR}, 2018.

\bibitem{xu2015show}
K.~Xu, J.~Ba, R.~Kiros, K.~Cho, A.~Courville, R.~Salakhudinov, R.~Zemel, and
  Y.~Bengio.
\newblock Show, attend and tell: Neural image caption generation with visual
  attention.
\newblock In {\em ICML}, 2015.

\bibitem{yao2015describing}
L.~Yao, A.~Torabi, K.~Cho, N.~Ballas, C.~Pal, H.~Larochelle, and A.~Courville.
\newblock Describing videos by exploiting temporal structure.
\newblock In {\em ICCV}, 2015.

\bibitem{yeung2018every}
S.~Yeung, O.~Russakovsky, N.~Jin, M.~Andriluka, G.~Mori, and L.~Fei-Fei.
\newblock Every moment counts: Dense detailed labeling of actions in complex
  videos.
\newblock {\em International Journal of Computer Vision (IJCV)}, 2018.

\bibitem{yeung2016glimpses}
S.~Yeung, O.~Russakovsky, G.~Mori, and L.~Fei-Fei.
\newblock End-to-end learning of action detection from frame glimpses in
  videos.
\newblock In {\em CVPR}, 2016.

\end{thebibliography}
}

\end{document}